\documentclass[sigconf]{acmart}
\usepackage{amsmath,amssymb, graphicx}
\usepackage[ruled,linesnumbered]{algorithm2e}
\usepackage{algorithmic}
\usepackage{paralist} 
\usepackage{color}
\usepackage{multirow}
\usepackage{rotating}
\usepackage{hyperref}
\usepackage[mathscr]{eucal} 
\usepackage{amsbsy} 
\usepackage{booktabs}
\usepackage{xcolor}
\usepackage{tikz}
\usetikzlibrary{snakes}
\usepackage{multirow}
\usepackage{epstopdf}
\usepackage{balance}
\usepackage{tablefootnote}
\usepackage{empheq} 
\usepackage{moresize}
\usepackage{enumitem}
\usepackage{subcaption}
\setlist{nolistsep}

\usepackage[font=small,labelfont=bf,skip=0pt]{caption}

\DeclareCaptionType{copyrightbox}
\usepackage{url}
\usepackage{etoolbox}

\newcounter{ALC@tempcntr}
\newcommand{\tensor}[1]{\underline{ \mathbf{#1} }}
\newcommand{\M}[1]{\mathbf{#1}} 
\newcommand{\shield}{\textsc{Shield}\xspace}

\newcommand{\reminder}[1]{{\textsf{\textcolor{red}  {[#1]}}}}
\newcommand{\hide}[1]{}

\newcommand{\ben}{\begin{enumerate*}}
\newcommand{\een}{\end{enumerate*}}
\newcommand{\bit}{\begin{itemize*}}
\newcommand{\eit}{\end{itemize*}}

\hide{

}

\makeatletter
\def\@copyrightspace{\relax}
\makeatother
\setcopyright{none} 
\acmConference[]{ }{ }{ }

\acmYear{}
\copyrightyear{}
\acmPrice{}
\acmISBN{}
\acmDOI{}
\makeatother

\setcopyright{none}
\renewcommand\footnotetextcopyrightpermission[1]{}
\begin{abstract}
Recent studies have demonstrated that machine learning approaches like deep neural networks (DNNs) are easily fooled by adversarial attacks. Subtle and imperceptible perturbations of the data are able to change the result of deep neural networks. Leveraging vulnerable machine learning methods raises many concerns especially in domains where security is an important factor. Therefore, it is crucial to design defense mechanisms against adversarial attacks. For the task of image classification, unnoticeable perturbations mostly occur in the high-frequency spectrum of the image. In this paper, we utilize tensor decomposition techniques as a preprocessing step to find a low-rank approximation of images which can significantly discard high-frequency perturbations. Recently a defense framework called \shield \cite{das2018shield} could ``vaccinate" Convolutional Neural Networks (CNN) against adversarial examples by performing random-quality JPEG compressions on local patches of images on the ImageNet dataset. Our tensor-based defense mechanism outperforms the SLQ method from \shield by 14\% against Fast Gradient Descent (FGSM) adversarial attacks, while maintaining comparable speed.
\end{abstract}

\begin{document}
\author{Negin Entezari}
\affiliation{{University of California Riverside}}
\email{nente001@ucr.edu}

\author{Evangelos E. Papalexakis}
\affiliation{{University of California Riverside}}
\email{epapalex@cs.ucr.edu}
\title{\textsc{TensorShield}: Tensor-based Defense Against Adversarial Attacks on Images}
\maketitle

\section{Introduction}
\label{sec:intro}
In the last few year, Deep Neural Networks (DNNs) have been tremendously popular in a various domains including image processing and computer vision. However, recently, the robustness of DNNs has been questioned when facing adversarial inputs. The performance of DNNs can significantly drop even on slightly  perturbed instances \cite{szegedy2013intriguing}. For the task of image classification, attackers put constraints on perturbations such that they remain unnoticeable to the human eye, but they are still able to greatly deteriorate the performance of the model. 

Utilizing machine learning methods which are vulnerable to adversarial attacks in system where safety and security are critical factors may cause serious problems. Therefore, it is crucial to have a robust model against adversaries, especially in security-sensitive domains like autonomous driving and medical imaging. To address this concern, recent studies have conducted research to analyze vulnerability of deep learning methods in order to come up with defense techniques against the adversarial attacks \cite{das2018shield,bhagoji2017dimensionality,metzen2017detecting,papernot2016distillation}. 

To measure the strength of a perturbation, usually an $l_2$ or $l_\infty$ norm is used. Adversarial perturbations are mostly designed so that they have a small norm and are unnoticeable to human inspection. Designing a defense mechanism is a difficult task. Typically, the defender has only access to the perturbed instances (and definitely not the original ones, where there would be hope to identify which parts have been tampered with) and should be able to defend against different types of perturbations. Moreover, a defense mechanism which specialized on a particular kind of attack could be easily defeated by new attacks which are optimized against its strategy. Therefore, designing a  defense technique which captures a universal pattern across various attacks is highly desirable, since this will  able to defend against most of the adversarial attacks. 

\shield proposed by Das et al. \cite{das2018shield}, is a real-time defense framework which performs JPEG compression with random levels over local patches of images to eliminate unnoticeable perturbations which mostly appear in high frequency spectrum of images. In this paper, we propose a tensor decomposition approach to compute a low-rank approximation of images which significantly discards high-rank perturbations. However, \shield considers images in isolation and does not pay attention to the correlation of images when facing adversarial attacks. \hide{Moreover, the final images after performing the proposed tensor decomposition defense mechanism, look more realistic and less blurry compared to those resulting from \shield.} 

Our contributions are as follows:
\begin{itemize}[noitemsep]
	\item {\bf Defense through the lens of factorization}: We propose a novel defense against adversarial attacks on images which utilizes tensor decomposition to reconstruct a low-approximation of perturbed images before feeding them to the deep network for classification. Without any retraining of the model, our method can significantly mitigate adversarial attacks.
	\item {\bf Efficient and effective method}: Representing images with tensor, allows processing images in batches as 4-mode tensor, which is able to capture latent structure of perturbations from multiple images rather than a single image which leads to more performance improvements. 
\end{itemize}
The rest of this paper is organized as follows. In Section \ref{sec:related} we discuss related work. We introduce our proposed method in Section \ref{sec:method} and provide experimental results in Section \ref{sec:experiments}. Finally, in section \ref{sec:conclusions} we offer conclusions and discuss future works.

\begin{figure}
\centering
\includegraphics[width=0.99\linewidth]{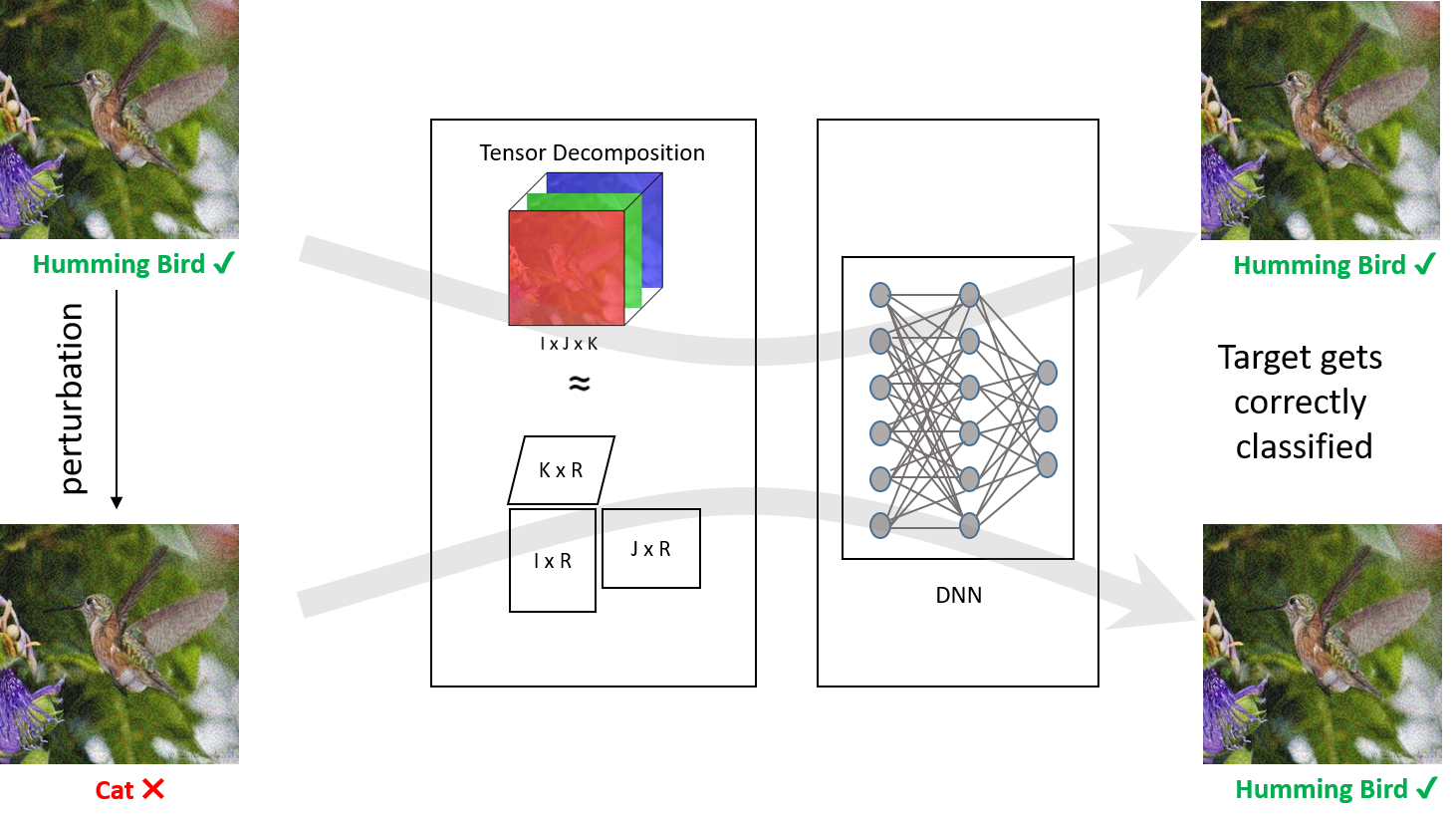}
\caption{System Overview: low-rank tensor approximation of images to ``vaccinate" the network against perturbations. (the term ``vaccinate'' was first used by Das et al. \cite{das2018shield} to refer to models equipped with a defense mechanism.)}
\label{fig:systemoverview}
\end{figure}

\hide{
Our contributions include:
\begin{itemize}[noitemsep]
	\item {\bf Contribution 1}: Brief description
	\item {\bf Contribution 2}: Brief description
	\item {\bf Contribution 3}: Brief description
\end{itemize}
}
\section{Related Work}
\label{sec:related}
\subsection{Adversarial Attacks}
In this paper, we focus on defending against adversarial attacks on deep learning methods for the task of image classification. Here, we briefly outline some of the most popular adversarial attacks on images.

Given a classifier $C$, the goal of an adversarial attack is to modify an instance $x$ to a perturbed instance $x'$ such that $C(x) \neq C(x')$, while keeping the distance $\|x-x'\|$ between perturbed and clean instance small. By $\|.\|$ we deonte some norm which is also used to express the strength of the perturbations. The popular choices are Euclidean distance ($l_2$ norm) and Chebyshev distance ($l_\infty$ norm). Here, we discuss some of the popular attacks, against which we evaluate our proposed method. 

\textbf{Fast Gradient Sign Method (FGSM)}\cite{goodfellow2014explaining}:  FGSM is a fast method to compute perturbations which is based on computing first-order gradients. FGSM generates adversarial images by introducing a perturbation as follows:
\begin{equation}
    x'=x+\epsilon.sign(\nabla J_x(\theta,x,y))
\end{equation}
where $\epsilon$ is a user-defined threshold  that determines the strength of the perturbations and controls the magnitude of perturbations per pixel. $\theta$ is the parameter of the model, $y$ is the true label of the instance $x$, and $J$ is the cost of training the neural network.

\textbf{Iterative Fast Gradient Sign Method (I-FGSM)}\cite{kurakin2016adversarial}: I-FGSM is the iterative version of the FGSM. In each iteration $i$, I-FGSM clips the pixel values to remain within the $l_\infty$ neighborhood of the corresponding values from a ``clean'' instance $x$:
\begin{equation}
    x'_i=x'_{i-1}+\alpha.sign(\nabla J_{x'_{i-1}}(\theta,x'_{i-1},y))
\end{equation}

\textbf{Projected Gradient Descent (PGD)}\cite{madry2017towards}:
PGD is one of the strongest gradient-based attacks \cite{madry2017towards} 
Given a clean image $x$, PGD aims to find a small perturbation $\delta \in \mathcal{S}$ to generate the perturbed instance $x' = x+\delta$. PGD starts from a random perturbation and iteratively updates the perturbation:
\begin{equation}
    \delta_i =\Pi_\mathcal{S}[\delta_{i-1}+\tau sign(\nabla_x L(x+\delta_{i-1},y))]
\end{equation}
where $\tau$ is a fixed step size. $\Pi_\mathcal{S}$ projects the perturbation onto set $\mathcal{S}$, set of allowed perturbation in the $\epsilon$ neighborhood the ``clean" instance $x$.

\subsection{Defense Against Adversarial Attacks}
\shield proposed by Das et al. \cite{das2018shield}, uses image preprocessing as a defense mechanism to reduce the effect of perturbations. \shield is based on the observation that the attacks described above are high-frequency, thus, eliminating those high frequencies (which are not generally visible by the human eye) will sanitize the image. \shield performs Stochastic Local Quantization (SLQ) as a preprocessing step and subsequently employs JPEG compression with qualities 20, 40, 60, and 80 on the image, then for each $8 \times 8 $ block of the image, randomly selects from one of the compressed images. \shield also retrains the model on images compressed with different JPEG qualities and uses an ensemble of these models to defends against adversarial attacks. 

In this paper, we preprocess images using tensor decomposition techniques to achieve a low-rank approximation of the image. We can significantly alleviate the effect of perturbations without performing any retraining. In a parallel approach \cite{entezari2020} employs singular value decomposition to compute a low-rank approximation of graph to defend against adversarial attacks on graphs. However, this paper is the first to identify and leverage the observation that gradient-based attacks on deep learning image classifiers are manifested in high-rank components of a decomposition of the image.

\section{Proposed Method}
\label{sec:method}
In this section, we first investigate the characteristics of adversarial attacks on networks designed for the task of image classification. Then we propose a tensor-based defense mechanism against these attacks which improves the performance of the network.

\subsection{Characteristics of Image Perturbations }
Assume a trained model $C$ with a high accuracy on clean images is given. Adversarial attacks perform perturbations on the clean images in a way that they are imperceptible to humans, yet are successful in deceiving the model to misclassify the perturbed instances. In other words, for a clean image $x$ and its corresponding perturbed image $x'$, the goal is to have: $C(x) \neq C(x')$. The adversarial attacks do not preserve the spectral characteristics of images and add high frequency components to images to remain unnoticeable to the human eyes \cite{das2018shield}. Perturbations in image domain are crafted in a way that mostly affect high frequency spectrum of images. Therefore, discarding the high frequency factors of the image using approaches like compression or low-rank approximation of images could be successful defense mechanisms against these type of perturbations. Therefore, a mechanism that only keeps the low-rank components of the image and discards the high-rank ones, can
be successful in discarding the perturbations. In \cite{das2018shield}, authors leverage JPEG compression to remove high frequency components of the image and alleviate the effect of perturbations. In this paper, we study the problem from a ``matrix spectrum'' point of view (i.e., the singular value profile and the intrinsic low-rank dimensionality of the data) and use tensor decomposition techniques to achieve a low-rank approximation of perturbed images.
\hide{
RGB images can be modeled using tensors and then utilizing tensor decomposition methods we can achieve a low-rank approximation of the image that helps to get rid of high frequency components of the image.
}

\hide{One might think that this is not a valid claim \reminder{ref: A Fourier Perspective on Model Robustness in Computer Vision}. }

\subsection{Tensor-based Defense Mechanism}
In this section, we briefly describe concepts and notations used in the paper.

A tensor, denoted by $\tensor{X}$, is a multidimensional matrix. The order of a tensor is the number of modes/ways and is the number of indices required to index the tensor \cite{papalexakis2017tensors}. An RGB image is a three-mode tensor where the first and second modes correspond to the pixels and the third mode corresponds to the red, green, and blue channels, i.e. the frontal slices are red, green, and blue channels of the image. An RGB image of size $W \times H$ is a 3-mode tensor of size $W \times H \times 3$, where $W$ and $H$ are width and height of the image, respectively.

To achieve a low-rank approximation of the perturbed images, we perform a tensor decomposition technique on the image and by choosing small values for the rank of the tensor, we reconstruct a low-rank approximation of the image which is fed to the deep network. The low-rank approximation of image discards high frequency perturbations which can improve the performance of the network on the perturbed images. However, traditional tensor decomposition techniques like CP/Parafac \cite{parafac} and Tucker\cite{tucker3} are time-consuming and may slow down the neural network performance which makes our proposed method impractical for real-time defense. To overcome this issue, we leverage Tensor-Train decomposition \cite{oseledets2011tensor} which scales linearly with respect to the dimension of the tensor and was especially introduced to address the problem of curse of dimensionality \cite{oseledets2011tensor}. This highly-desirable property of the Tensor-Train allows us to process images in batches which form a 4-mode tensor and perform the Tensor-Train decomposition on 4-mode tensors quite fast. For a batch of $N$ images, the size of the 4-mode tensor will be $N\times W\times H \times 3$. Generally decomposing a 4-mode tensor is slower compared to a 3-mode one, however, by considering images in batches, some of the I/O overhead is reduced which results in almost the same processing time on the entire dataset.  Furthermore, processing images in batches improves the performance of the model. The reason behind this is that decomposing images in batches, extracts latent structure corresponding to perturbations from multiple images and captures general characteristics of perturbations. 

For a 4-mode tensor, the Tensor-Train decomposition can be written as follows:
\begin{equation}
    \label{eq:tensortrain}
    \tensor{X}(i,j,k,l) \approx \sum_{r_1,r_2,r_3} \mathbf{G}_1(i,r_1) \tensor{G}_2(r_1,j,r_2) \tensor{G}_3(r_2,k,r_3) \mathbf{G}_4(r_3,l)  
\end{equation}

\begin{figure}[t]
\centering
\includegraphics[width=0.9\linewidth]{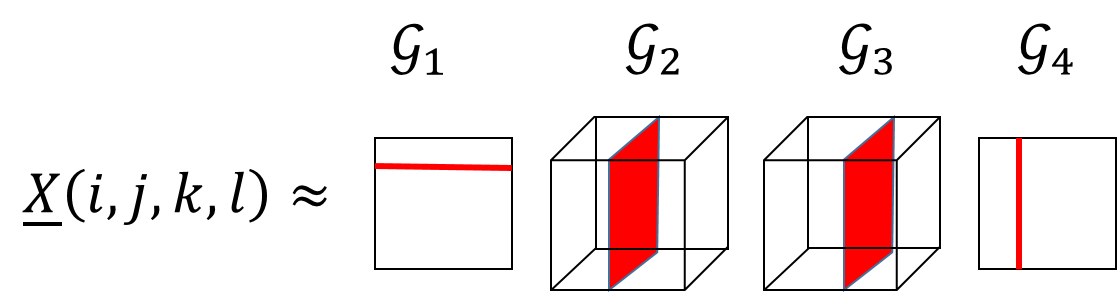}
\caption{Tensor-Train decomposition of a 4-mode tensor.}
\label{fig:TensorTrain}
\end{figure}

Figure \ref{fig:TensorTrain} illustrates the Tensor-Train decomposition of a 4-mode tensor.

\begin{figure}[b]
\centering
\includegraphics[width=0.9\linewidth]{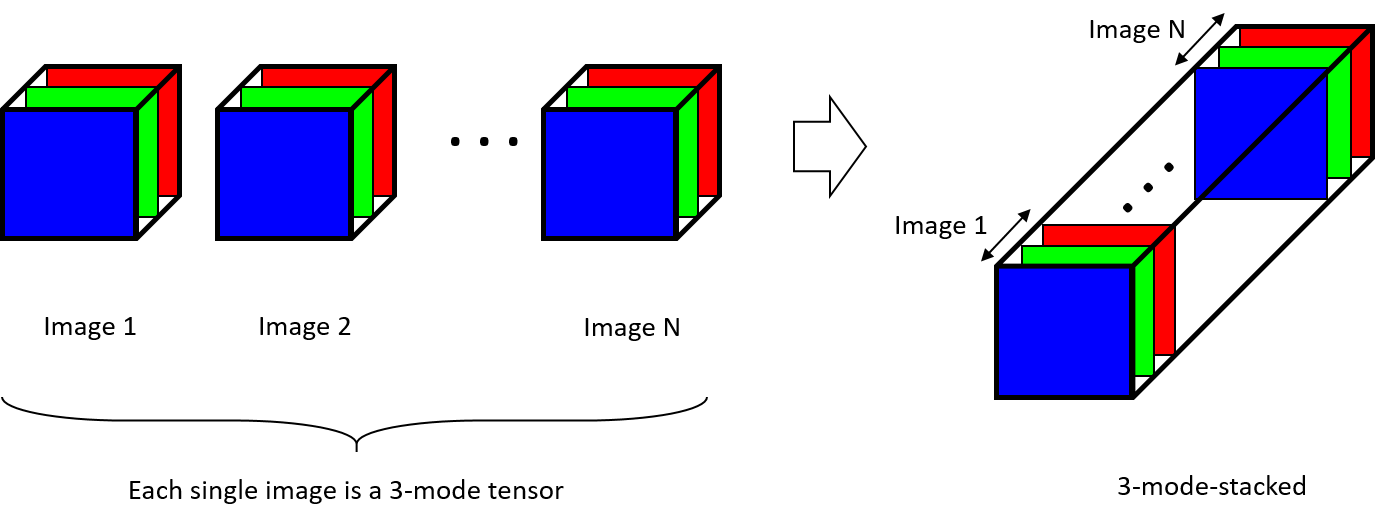}
\caption{Stacking 3-mode images along the third mode.}
\label{fig:stacked}
\end{figure}

Another possible representation for the batch of images is to convert the 4-mode tensor to a 3-mode tensor by stacking the images along the third mode, i.e. stacking RGB channels and the result tensor will be of dimension $W \times H \times 3*N$. Figure \ref{fig:stacked} illustrates a 3-mode stacked tensor of $N$ images. 
There are other ways to convert a 4-mode tensor into a 3-mode one. For instance, another way is to flatten the RGB image into a matrix with three columns corresponding to the channels of the image. With this representation, the final tensor will be of size $W*H \times 3 \times N$. One disadvantage of this representation is that flattening the image ignores the spatial relationship of the pixels. Moreover, with this vectorized representation, the first dimension is much bigger than the other two dimensions and requires a larger value of rank to get a reasonable approximation of the image, and larger ranks make the decomposition slower. For these reasons, we do not consider the vectorized representation in our study. In the experimental evaluations that follows, we will examine different representation including single image versus batch of images and 3-mode tensors versus 4-mode tensors.

\hide{
Images in the ImageNet dataset are of size $299 \times 299$, therefore, the tensor size is $299\times 299 \times 3$. By selecting small values for $R_1$, $R_2$, and $R_3$ we can achieve a low-rank approximation of images and discard the high-frequency perturbations. Therefore, for each image in the dataset, we perform a tucker decomposition and the resulting low-ran approximation of the image is fed to the network for the task of classification. With this approach we could greatly defend perturbations and our results was better than those of \shield. In the experimental evaluations that follows, We will compare the results in details.

However, performing tucker decomposition per image is very slow. To overcome this issue, we consider images in batches of size $N$. This leads to a four-mode tensor of size $N\times 299 \times 299 \times 3$. With $N = 30$, the process got about five times faster than before but still five times slower than \shield. We observed that decomposing images in batches improved the performance of the model by almost 2\%. The reason behind this is that decomposing images in batches, considers the correlation between images and better captures the characteristics of adversarial perturbations. To gain more speedups, we convert the four-mode tensor to a three-mode tensor. An important factor impacting the speed of the decomposition is the decomposition ranks $(R_1,R_2,R_3)$. Decomposition at lower ranks is significantly faster than higher ranks. And a factor impacting the ranks is the size of the corresponding dimension. Let assume we flatten the images into a vector of size $299*299(=89401)$ per channel. So the final tensor will be of size $N\times 89401 \times 3$. In this case, $R_2$ needs to be much larger than $R_1$ and $R_3$ to achieve a proper approximation of images. We observed that if the tensor is balanced i.e. the dimensions are almost equal, decomposition speed increases. Stacking images along the third dimension results in more balanced tensors. We convert the four-mode tensor to three-mode by stacking images on the third dimension resulting to a $299 \times 299 \times 3*N$. We will discuss the relationship between ranks, accuracy and speed in experimental evaluations.
}
\hide{
Among a variety of tensor decomposition techniques, Tucker decomposition is one of the most popular methods. Given a three-mode tensor $X \in \mathbb{R}^{I \times J \times K}$ its tucker decomposition \reminder{ref} is defined as follows:
\begin{equation}
\label{eq:tucker2}
    \tensor{X} \approx \mathscr{G} \times_1 \M{U_1} \times_2 \M{U_2} \times_3 \M{U_3}
\end{equation}
where $\mathscr{G}\in \mathbb{R}^{R_1 \times R_2 \times R_3}$ is the core tensor, $\M{U_1} \in \mathbb{R}^{I\times R_1}$, $\M{U_2} \in \mathbb{R}^{J\times R_2}$, and $\M{U_3} \in \mathbb{R}^{K\times R_3}$ are factor matrices. The tucker decomposition can be also written as:
\begin{equation}
\label{eq:tucker2}
\begin{split}
    &\tensor{X}(i,j,k) \approx \\ &\sum_{r_1=1}^{R_1}\sum_{r_2=1}^{R_2}\sum_{r_3=1}^{R_3} \mathscr{G}(r_1,r_2,r_3)\M{U_1}(i,r_1)\M{U_2}(j,r_2)\M{U_3}(k,r_3)
    \end{split}
\end{equation}

The core matrix $\mathscr{G}$ shows interactions between factor matrices and when $R_1 < I$, $R_2 < J$, and $R_3 < K$, it is considered as a compression of tensor $\tensor{X}$. For more details about tensors and tensor decomposition, we refer the interested reader to \cite{papalexakis2017tensors, kolda2009tensor}.
}

\section{Experimental Evaluation}
\label{sec:experiments}
In this section, we show how the proposed method can successfully remove adversarial perturbations and we compare our results to \shield (SLQ). According to \cite{cornelius2019efficacy}, original \shield evaluations has gained benefit from central cropping of images during evaluation, whereas the perturbations were generated with cropping being off. In all our evaluations, we disable the central cropping.
\subsection{Experiment Setup}
We performed experiments on the validation set of the \textit{ImageNet} dataset which includes 50,000 images from 1,000 classes. All experiments are performed on the \textit{ResNet-v2 50} model from the \textit{TF-Slim} module of \textit{TensorFlow}. The adversarial attacks are from the CleverHans package \footnote{https://github.com/tensorflow/cleverhans} \cite{papernot2016cleverhans}. We performed the experiments on a machine with one NVIDIA Titan Xp (12 GB) GPU. We used TensorLy \footnote{https://github.com/tensorly/tensorly} library in Python to perform tensor decomposition techniques \cite{kossaifi2019tensorly}. 

\subsection{Parameter Tuning}
In our evaluations, we express different configurations in form of a list as  [tensor decomposition, tensor representation, batch size, rank] and we investigate the accuracy and runtime of the ResNet-v2 50 on 1000 images from the ImageNet dataset for different configurations. The possible values for each part of the configuration list is as follows:
\begin{itemize}
    \item Tensor decomposition: \{Parafac, Tucker, Tensor-Train\}
    \item Tensor representation: \{3-mode, 3-mode-stacked, 4-mode\}
    \item Batch size: \{1, 5, 10, 20, 50\}
    \item Rank: varies by choice of tensor representation and decomposition.
\end{itemize}

\hide{
\begin{enumerate}
    \item{{Tensor representation}}:
    \begin{itemize}
        \item Single image tensor (3-mode) i.e. batch size is one.
        \item Stacked batch of images (3-mode).
        \item Batch of images (4-mode).
    \end{itemize}
    \item{{Tensor decomposition method and its ranks}}:
    \begin{itemize}
        \item Parafac
        \item Tucker
        \item Tensor-Train
    \end{itemize}
\end{enumerate}
}

\begin{figure}[t]
\centering
\begin{minipage}{0.9\linewidth}
\centering
\includegraphics[width=\linewidth]{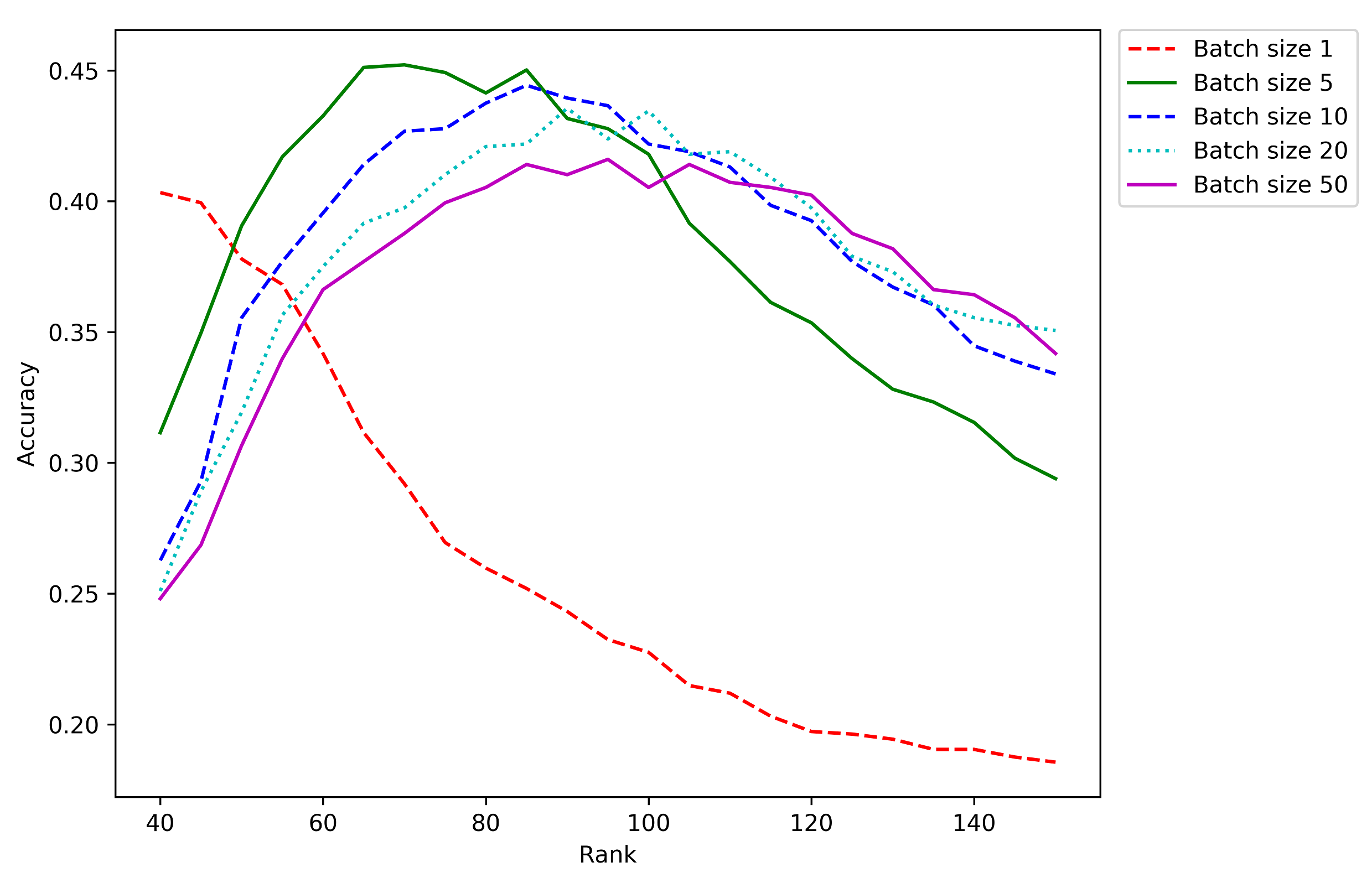}
\subcaption{Accuracy} \label{fig:TT_accuracy}
\end{minipage}
\begin{minipage}{0.9\linewidth}
\centering
\includegraphics[width=\linewidth]{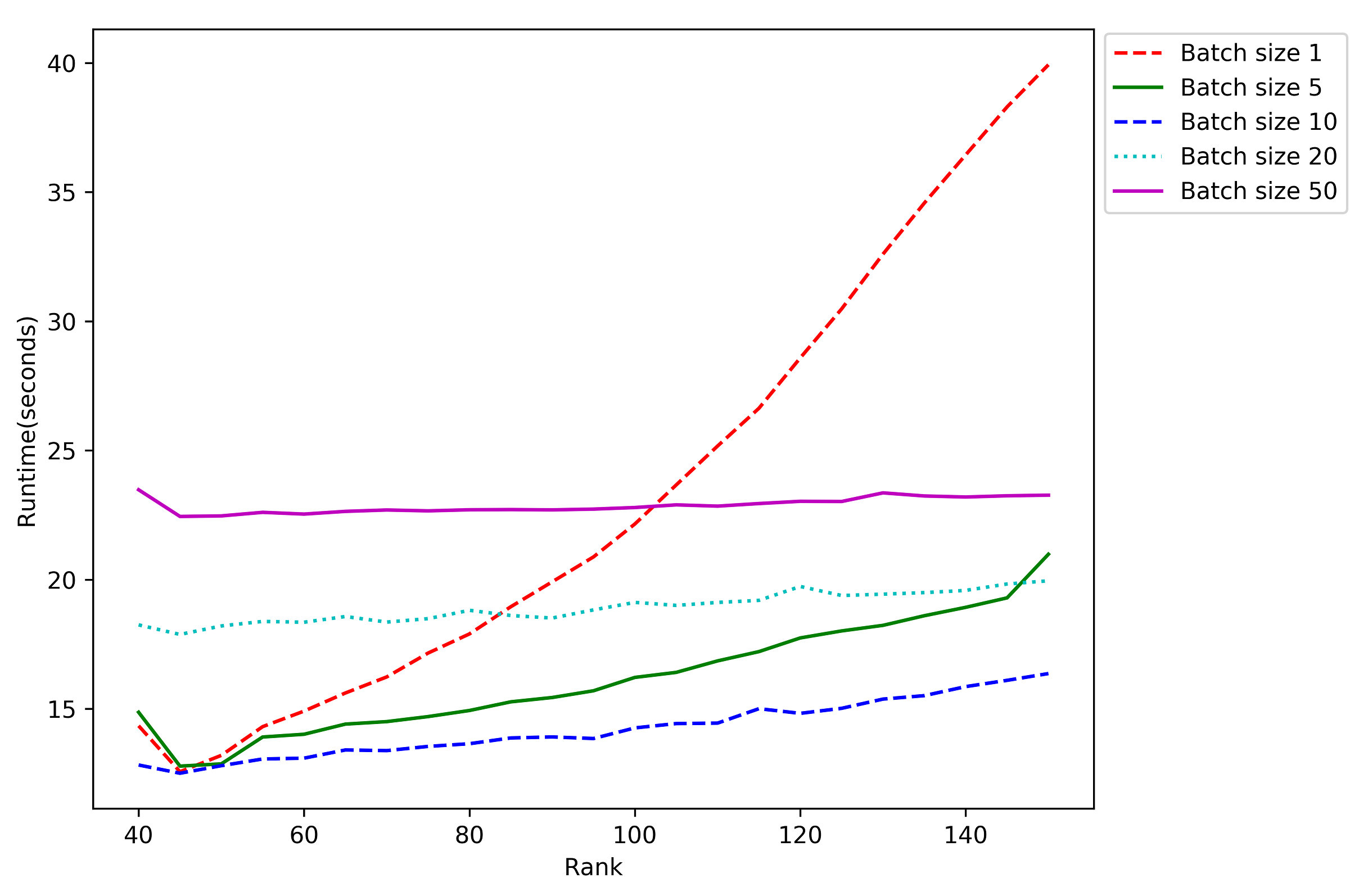}
\subcaption{Runtime} \label{fig:TT_runtime}
\end{minipage}
\vspace{0.2cm}
\caption{Accuracy and runtime of ResNet-v2 50 over 1000 images attacked by FGSM ($\epsilon = 4$). Tensor-Train decomposition is applied on a single image (batch size 1) or \textbf{4-mode} tensor of batches of size 5, 10, 20, and 50 to defend against FGSM perturbations.}
\label{fig:TT_batchsize}
\end{figure}

Performing tensor decomposition for a batch of images can reduce the decomposition overhead compared to decomposing a single image and accelerates the entire evaluation process. Moreover, considering images in batches helps to better capture the pattern of perturbations from multiple images. However, the choice of the right batch size is important. A large batch of images needs larger ranks for decomposition and could get very slow. Also, in a large batch of images, the variety of images which are from different classes increases which deteriorates the performance of the decomposition. To find the best batch size, we perform a grid search on values 5, 10, 20, and 50. Tensor Train decomposition of a 4-mode tensor requires setting 3 values for the ranks. The first value corresponds to compressing the batches, the second value corresponds to compressing the image pixels, and the third value corresponds to compressing the RGB channels. We fix the first rank to the number of batches and the third rank to the number of channels i.e. 3. For the second rank, we search within range 40 to 150. Figure \ref{fig:TT_batchsize} shows the accuracy and runtime of the model for different batch sizes for Tensor-Train decomposition with ranks ranging from 50 to 120 with steps of 5. The figure also shows how processing single images (batch size 1) differs from batch sizes greater than 5. In the case that we are processing single images, the runtime increases as the rank gets larger, however, as the batch size increase, the runtime becomes less sensitive to the ranks and for the batch size 50 it will become almost constant for all the ranks. Batch size 5 produces the highest accuracy, while batch size 10 has the lowest runtime. There is a trade-off between runtime and accuracy. Based on the priorities of the system, one might sacrifice accuracy for speed. 

\begin{figure}[t]
\centering
\begin{minipage}{0.9\linewidth}
\centering
\includegraphics[width=\linewidth]{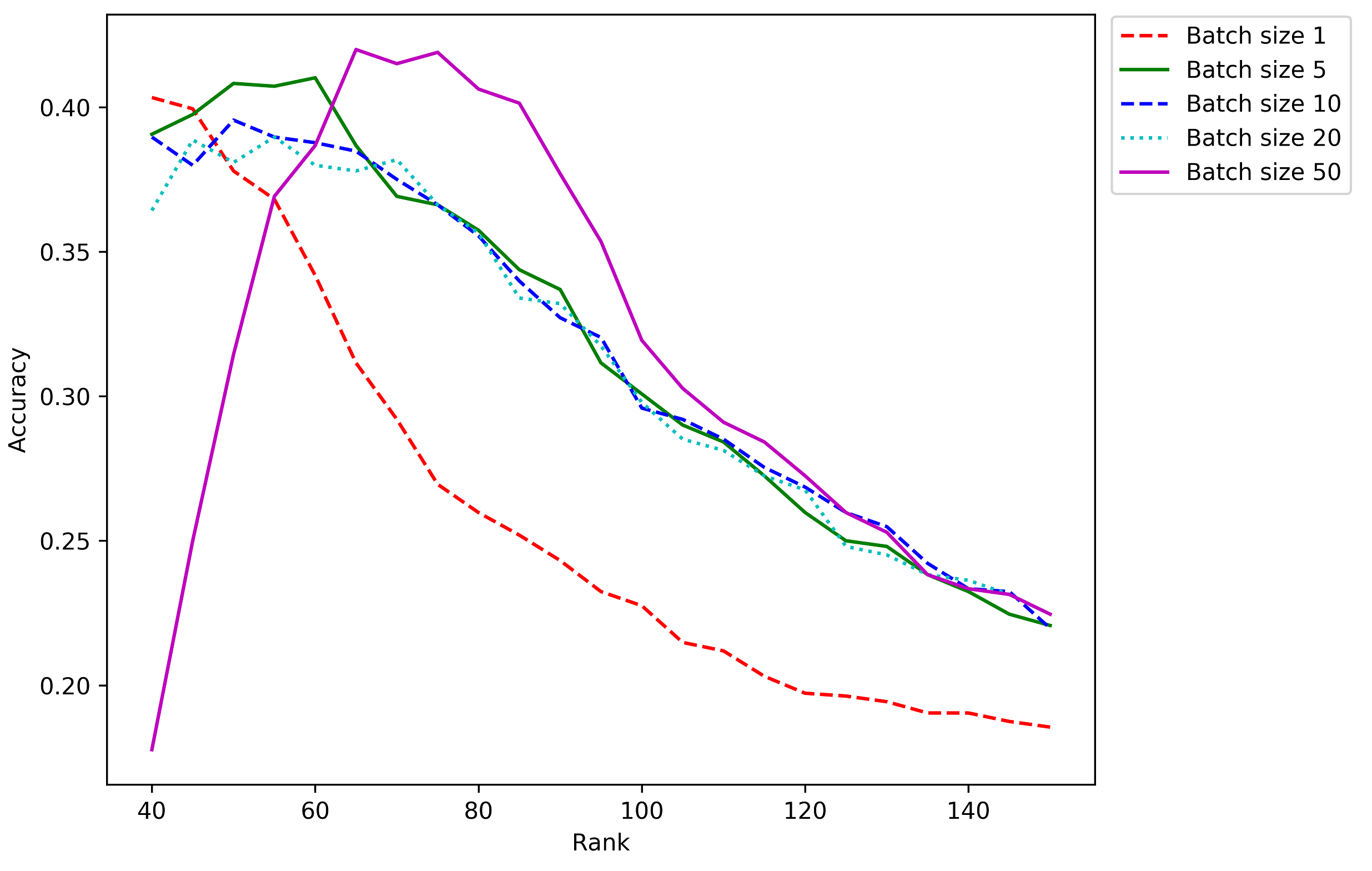}
\subcaption{Accuracy} \label{fig:TT_stacked_accuracy}
\end{minipage}
\begin{minipage}{0.9\linewidth}
\centering
\includegraphics[width=\linewidth]{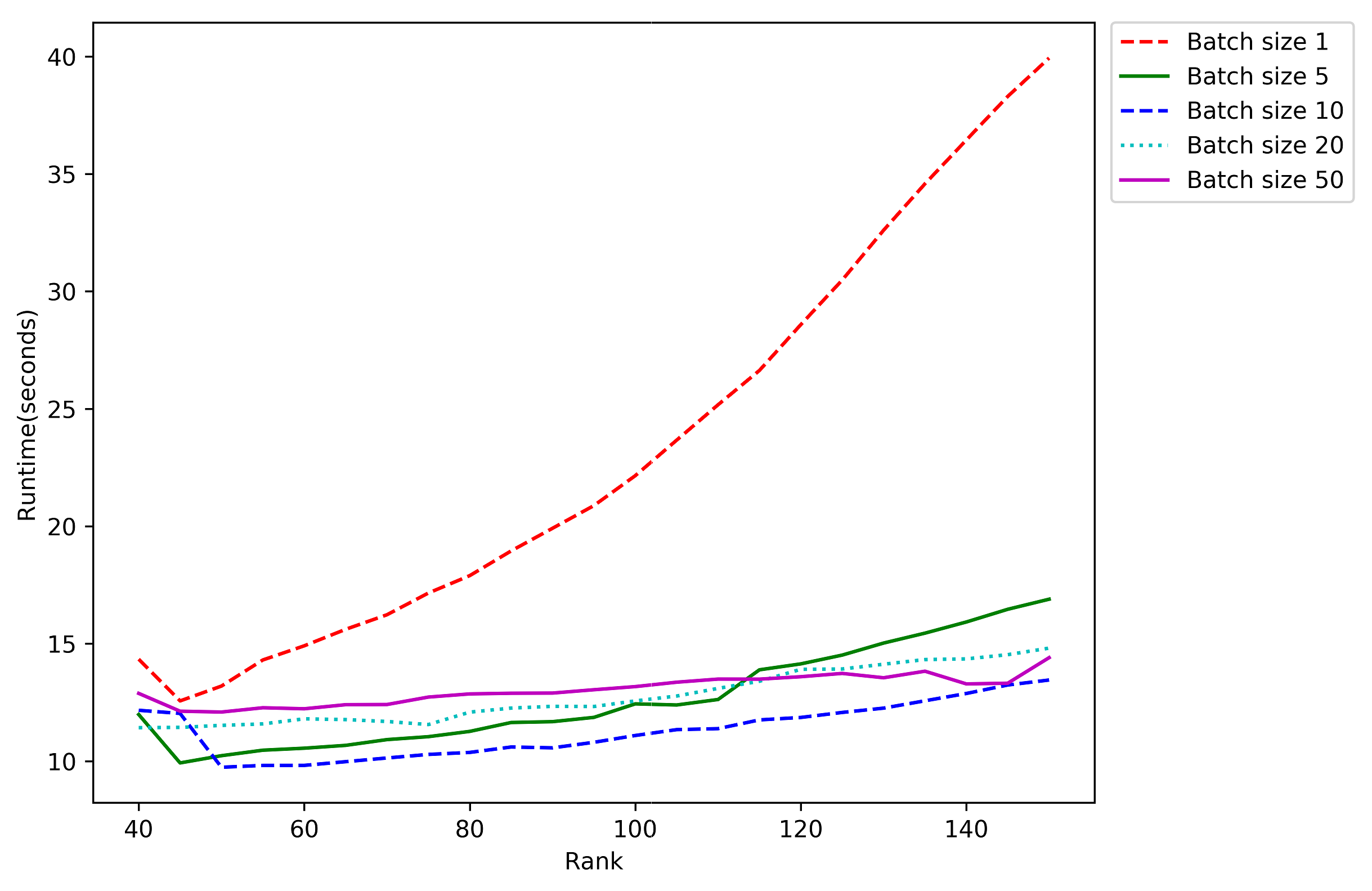}
\subcaption{Runtime} \label{fig:TT_stacked_runtime}
\end{minipage}
\vspace{0.2cm}
\caption{Accuracy and runtime of ResNet-v2 50 over 1000 images attacked by FGSM ($\epsilon = 4$). Tensor-Train decomposition is applied on a single image (batch size 1) or \textbf{3-mode-stacked} tensor of batches of size 5, 10, 20, and 50 to defend against FGSM perturbations.}
\label{fig:TT_batchsize_stacked}
\end{figure}

Figure \ref{fig:TT_batchsize_stacked} shows the effect of different batch sizes on the 3-mode-stacked representation. Plots for batch sizes 5, 10, and 20 are almost identical in both accuracy and runtime. Batch size 50 produces the highest accuracy with the 3-mode-stacked representation. However, the highest accuracy with 3-mode-stacked representation is lower than the highest accuracy achieved using the 4-mode representation.

\subsection{Results}
As mentioned in Section \ref{sec:method}, Tensor-Train performs much faster than Parafac and Tucker. Therefore, for the Parafac and Tucker, we only report the result for the configuration which corresponds to the maximum accuracy, as a reference for comparison against Tensor-Train. Table \ref{tab:result} shows the result.

\begin{table*}[!hbt]
    \centering
        \begin{tabular}{l|ccc|c}
        \textbf{Configurations} & \textbf{PGD} & \textbf{FGSM } & \textbf{i-FGSM} & \textbf{Runtime}\\
         & ($\epsilon = 4$) & ($\epsilon = 4$) & ($\epsilon = 4$) & (seconds)\\
        \hline
        No defense & 11.10 & 18.40 & 7.49 & \\[0.5ex]
        [Tensor-Train, 4-mode, 5, [5,90,3]] & \textbf{51.53} & \textbf{43.59} & \textbf{50.46} & 675 \\ [0.5ex]
        [Tensor-Train, 4-mode, 10, [10,100,3]] & 51.01 & 43.10 & 49.95 & 605 \\ [0.5ex]
        [Tensor-Train, 3-mode, 1, 40] & 49.75 & 42.32 & 48.52 & \textbf{530} \\ [0.5ex] 
        [Tucker, 3-mode-stacked, 30, [105,105,90]] & 49.37 & 40.07 & 48.79& 1050 \\ [0.5ex]
        [Parafac, 3-mode, 1, 60] & 48.11 & 41.38 & 49.75 & 5500 \\ [0.5ex] 
        SLQ & 44.60 & 29.40 & 38.60 & 410 \\ [0.5ex] 
        \hline
        \end{tabular}
    \caption{Summary of accuracies and runtime of ResNet-v2 50 on ImageNet validation set against FGSM, i-FGSM, and PGD adversarial attacks for defenses with different configurations.}
    \label{tab:result}
\end{table*}

As illustrated in Table \ref{tab:result}, Tensor-Train outperforms Tucker and Parafac with respect to both accuracy and runtime. Tensor-Train performed on 4-mode tensor has produces the highest accuracy. As explained earlier, processing images in batches better captures latent components corresponding to perturbation by leveraging higher-order correlations. Tensor-Train can be utilized with different tensor representations (3-mode, 3-mode-stacked, or 4-mode) to adjust to needs for higher accuracy or higher speed. While the 4-mode representation produces the highest accuracy, the 3-mode single image representation can be used to speed up the process, with small drop in the accuracy.
SLQ is the fastest among all defenses, but it has the lowest accuracy. 

\begin{table}[!hb]
    \centering
        \begin{tabular}{l|c|cc}
        \textbf{Patch size} & \textbf{   Ranks  } & \textbf{   Accuracy   } & \textbf{  Runtime (seconds)}\\
         
        \hline
                & [5,10,3] &  &  \\ [0.5ex]
        [50,50] & [5,20,3] & 48.35 & 1100 \\ [0.5ex]
                & [5,30,3] &  & \\ [0.5ex]
                & [5,70,3] &  &  \\ [0.5ex]
                \hline
                & [5,40,3] &  &  \\ [0.5ex]
        [150,150] & [5,50,3] & 50.96 & 765 \\ [0.5ex]
                & [5,60,3] &  & \\ [0.5ex]
                \hline
                & [5,70,3] &  &  \\ [0.5ex]
        No patching & [5,90,3] & 50.48 & 710 \\ [0.5ex]
                & [5,110,3] &  & \\ [0.5ex]
    \end{tabular}
    \caption{Accuracies and runtime of ResNet-v2 50 on ImageNet validation set against PGD adversarial attacks with $\epsilon = 4$ vaccinated using Tensor-Train with 4-mode tensor of batch size 5. Decomposition rank is randomly selected from a set of possible ranks. No patching is equivalent to full size image.}
    \label{tab:randomness}
\end{table}

\hide{
\begin{figure*}[!ht]
\centering
\begin{minipage}{\textwidth}
\centering
\includegraphics[width=\textwidth]{FIG/defended1.png}
\end{minipage}
\begin{minipage}{\textwidth}
\centering
\includegraphics[width=\textwidth]{FIG/defended2.png}
\end{minipage}
\begin{minipage}{\textwidth}
\centering
\includegraphics[width=\textwidth]{FIG/defended3.png}
\end{minipage}
\begin{minipage}{\textwidth}
\centering
\includegraphics[width=\textwidth]{FIG/defended4.png}
\end{minipage}
\begin{minipage}{\textwidth}
\centering
\includegraphics[width=\textwidth]{FIG/defended5.png}
\end{minipage}
\caption{Adversarial image (attacked using PGD with $\epsilon = 4)$ is reconstructed using Tensor-Train on 4-mode tensors with batch size of 5 with rank [5,90,3] and Tensor-Train on 3-mode tensor of single images with rank 40. 
}
\label{fig:defended_image}
\end{figure*}
}

\subsection {Introducing Randomness to the Defense Framework}
Incorporating some randomness in the defense framework has makes the job of the attacker more difficult to deal with a random strategy rather than a fixed one. By selecting randomly from a set of ranks, we can add randomness to the tensor decomposition process. Another way is to split image into small patches, similar to local $8 \times 8$ patches from \shield, and perform decomposition of random rank on each patch and stitch up the patches to reconstruct a randomized low-rank approximation of images. In a 4-mode tensor representation, splitting images into patches creates smaller 4-mode tensors, e.g. splitting a 4-mode tensor including 5 batches of images with size $300 \times 300 \times 3$ into patches of size $50 \times 50$ creates 6 tensors of size $5,50,50,3$. Table \ref{tab:randomness} shows the results of incorporating randomness with tensor decomposition.  

\section{Conclusions}
\label{sec:conclusions}

In this paper, we explored to what extent low-rank tensor decomposition of perturbed images during the preprocessing step helps to defend against adversarial
attacks. The low-rank approximation of the perturbed image is then fed to the deep network for the task of classification. We evaluated our method against popular adversarial attacks: FGSM, I-FGSM, and PGD. We illustrated that considering images in small batches better captures the latent structure of perturbations and helps to improve the performance of the model. We also showed that how different configurations allow to trade-off between accuracy and runtime. 

\hide{
\section{Acknowledgements}
{\scriptsize
Research was supported by the National Science Foundation Grant No. XXXXXX. Any opinions, findings, and conclusions or recommendations expressed in this material are those of the author(s) and do not necessarily reflect the views of the funding parties.
}
}

\balance
\bibliographystyle{ACM-Reference-Format}
\bibliography{ms.bbl}

\end{document}